\documentclass[10pt,twocolumn,letterpaper]{article}

\usepackage{3dv}
\usepackage{times}
\usepackage{epsfig}
\usepackage{graphicx}
\usepackage{amsmath}
\usepackage{amssymb}

\usepackage[pagebackref=true,breaklinks=true,letterpaper=true,colorlinks,bookmarks=false]{hyperref}

\threedvfinalcopy % *** Uncomment this line for the final submission

\def\threedvPaperID{217} % *** Enter the 3DV Paper ID here

\ifthreedvfinal\fi
\begin{document}

%%%%%%%%% TITLE
\title{Do We Need Depth in\\ State-Of-The-Art Face Authentication?}

% \author{Amir Livne\inst{1} \and
% Alex Bronstein\inst{1,2} \and
% Ron Kimmel\inst{1,2} \and
% Ziv Aviv\inst{2} \and
% Shahaf Grofit\inst{2}}

% %
% \authorrunning{A. Livne \textit{et al.}}
% \institute{Technion, Israel Institute of Technology, Haifa, Israel\\ \email{\{alivne@campus,bron@cs,ron@cs\}.technion.ac.il}
% \and
% Intel EGI, Israel\\ \email{\{ziv.aviv,shahaf.grofit\}@intel.com}
% }

\author{Amir Livne$^1$\\
\and
Ziv Aviv$^{2}$\\
\and
Shahaf Grofit$^{2}$\\
\and
Alex Bronstein$^{1,2}$\\
\and
Ron Kimmel$^{1,2}$\\
\and
$^1$ Technion, Israel Institute of Technology, Haifa, Israel\\ \tt\small{\{alivne@campus,bron@cs,ron@cs\}.technion.ac.il}
\and
$^2$ Intel EGI, Israel\\ \tt\small{\{ziv.aviv,shahaf.grofit\}@intel.com}
}

\maketitle
% \thispagestyle{empty}

%%%%%%%%% ABSTRACT
\begin{abstract}
    Some face recognition methods are designed to utilize geometric information extracted from depth sensors to overcome the weaknesses of single-image based recognition technologies. 
    However, the accurate acquisition of the depth profile is an expensive and challenging process. 
    Here, we introduce a novel method that learns to recognize faces from stereo camera systems without the need to explicitly compute the facial surface or depth map. 
    The raw face stereo images along with the location in the image from which the face is extracted allow the proposed CNN to improve the recognition task while avoiding the need to explicitly handle the geometric structure of the face.
    This way, we keep the simplicity and cost efficiency of identity authentication from a single image, while enjoying the benefits of geometric data without explicitly reconstructing it.
    We demonstrate that the suggested method outperforms both existing single-image and explicit depth based methods on large-scale benchmarks, and even capable of recognize spoofing attacks.
    We also provide an ablation study that shows that the suggested method uses the face locations in the left and right images to encode informative features that improve the overall performance.
\end{abstract}

%%%%%%%%% BODY TEXT
\section{Introduction}
Automatic face recognition is a trusted biometric modality that often replaces passwords in modern smartphones and smart locks, payment applications, identity verification systems at border controls, etc. 
Common face recognition systems frequently use a single image to identify the subjects. 
When more than a single image is provided, {\it shape-from-X} techniques can be applied to extract the geometry of the observed object. 
This information is useful to overcome some of the acute challenges when considering a single-image based method, for example, handling extreme head poses, illumination variations, expressions \cite{singh2018techniques}, makeup, and spoofing attacks \cite{erdogmus2013spoofing}. 
Depth-based face recognition methods use the geometric structure of the face in order to gracefully handle these challenges, see \cite{bronstein2003expression,blanz2003face,paysan20093d} for early attempts in these directions. 
Furthermore, even without the texture image itself, the geometry alone is informative enough to recognize people \cite{bronstein2005expression,bronstein2006expression}. 
However, depth information is more complicated to acquire compared to traditional 2D (RGB) images, as it often requires costly sensors and calibration of the capturing system. 

The question we address in this paper is: ``can we design state-of-the-art face authentication systems that exploit geometric information without explicitly reconstructing the depth?". 
We suggest a novel approach that uses the geometric structure of the face to encode distinctive features for face recognition systems, without the need to explicitly reconstruct its surface. 
We accomplish this by feeding stereo images of a given face to a convolutional neural network (CNN) model together with the face locations in the \textit{left} and \textit{right} images. 
As long as the imaging system is fixed, there is no need to calibrate the cameras as nowhere in the processing pipeline do we compute the depth image or even rectify the images. 
With this input, the network has the information required to learn both geometric and photometric characteristics for a given face. 

Furthermore, we suggest a multi-task learning setting in which an auxiliary CNN is trained to receive the deep features from the face recognition CNN as the input, and construct a different viewpoint of the scene as the output. 
To accomplish this task, the deep features must represent some information regarding the geometric structure of the scene. 
Both the core face-recognition CNN and the auxiliary CNN are trained jointly, hence encouraging the face recognition model to encode geometric features in its latent space in an implicit manner. 
Since the auxiliary CNN is only used at training, it introduces no additional system costs at inference time.

We emphasize that the face recognition CNN, whether trained with or without the auxiliary CNN, does not require any explicit geometric data as an input. 
Also, the method is general and can be adapted to any CNN architecture. 
We show that though the \textit{left} and \textit{right} viewpoints are only slightly different, the proposed methods utilizes this information to significantly outperform single-image based methods on large-scale real stereo data.
To that end, we used state-of-the-art architectures as a baseline. 
Even more importantly, we tested our method on data generated with a 3D-morphable-model (3DMM) \cite{blanz1999morphable} that allows us to synthesize ground truth depth profiles, and show that the suggested method is more robust to head pose and illumination variability.
It also performs better than both traditional single-image based methods and a similarly-structured CNN, to which the ground-truth depth map is provided as an input. 
Furthermore, we show that the proposed method gracefully handles spoofing attacks, a challenging task in traditional 2D methods.

Our main contribution is a novel method to utilize geometric features from raw stereo data, without the need to explicitly reconstruct the depth or surface of the captured object. 
Our experiments show that in the domain of face recognition, using the suggested method makes explicit depth reconstruction superfluous at no performance cost. 
Consequently, it retains the strengths of 3D face recognition while enjoying the simplicity and cost-efficiency of existing 2D face recognition.

\section{Related Work}
\subsection{Computational face recognition} % an Open-Set Scenario}
Face recognition in an {\it open set scenario} refers to a pre-defined algorithm that provides a distance between two given images of faces, where the distance reflects the similarity between identities. 
An optimal algorithm should yield zero distance if the two images correspond to the same identity, and infinite distance otherwise. 
To that end, one often translates the given images into a more convenient representation space where they are referred to as feature vectors. 
Within the context of deep convolutional neural networks, these feature vectors are points in a latent space (also referred to as the \textit{embeddings} of the input), and their similarity can be evaluated in standard ways such as the Euclidean distance, correlation, or angle. 
This solution is suitable for the real-life challenge of face recognition, as it can be applied to unseen subjects, namely images of people that were not part of the training set.

Multiple loss functions by which the embedding is learned have been suggested over the years, such as softmax, constructive, and triplet loss.
Recent papers that report state-of-the-art results in that arena suggest angular margin losses \cite{liu2017sphereface,wang2018cosface,deng2019arcface} as a better measure of choice. 
In angular-loss based methods, during training time a linear classifier is trained to classify the different subjects based on their embedding vector. 
The training is typically done by minimizing the classification loss (e.g., cross-entropy)
\begin{equation}
    \mathcal{L}_{\mathrm{ang}} = \ell_{\mathrm{class}} ( h_{\mathrm{class}} ( h_{\mathrm{emb}}( f ) ), y  );
\end{equation}
here $h_{\mathrm{emb}}( f )$ denotes the embedding of the face image $f$, $y$ denotes the corresponding identity label, and $h_{\mathrm{class}}$ is a linear classifier predicting the label from the embedding vector. The loss is minimized with respect to the parameters of $h_{\mathrm{class}}$ and $h_{\mathrm{emb}}$. 

At inference time, the classifier is discarded and a similarity score is given by the angle (or the cosine of the angle) between the two embedding vectors
\begin{equation}
    S(f_1,f_2) = \frac {h_{\mathrm{emb}}(f_1)}{\left\|h_{\mathrm{emb}}(f_1)\right\|} \cdot \frac{h_{\mathrm{emb}}(f_2)}{\left\|h_{\mathrm{emb}}(f_2)\right\|};
\end{equation}

Most current methods apply a CNN to face images cropped from a  larger image of a scene (and then resized to match the model's input dimensions). 
The cropping procedure often exploits a facial landmarks detector, like the one reported in \cite{zhang2016joint}. 
When considering single image methods, the location of the face in the image is often ignored. 
When dealing with stereo images, the location of the face in the {\it left} and {\it right} images can be used to extract the depth profile of the face by using triangulation based methods. 
This additional geometric information can be useful for achieving better recognition rates. 

\subsection{Shape-from-stereo}
Stereo depth reconstruction methods try to evaluate the locations of objects in the 3D space given two images taken from slightly different viewpoints. 
Current practical stereo systems use two cameras, with a distance of several centimeters between their locations, also known as the camera baseline. 
By finding the projection of a specific point in the 3D space onto the two images (and the disparity between the projected coordinate in both images), it is possible to estimate the location of the point in 3D by back-projecting its location, a process known as {\it triangulation}. 
Though the method is theoretically simple, it is sensitive to the calibration of the stereo cameras and involves computationally intensive procedures to evaluate the exact correspondence between the two images, and the quality of the latter crucially depends on the image content.
Hence, recovering a good quality depth map (or surface) of an object becomes a challenging task in ``in-the-wild" scenarios. State-of-the-art systems often use active sensors with controlled illumination to overcome these difficulties, but these sensors are much more expensive than the passive triangulation-based cameras.

Recent efforts try to exploit the power of CNNs for object-specific depth reconstruction. 
Several papers had suggested methods to reconstruct the geometric surface of an object based on a single 2D-image. In particular, see the papers by Richardson et al. \cite{richardson2017learning,richardson20163d,sela2017unrestricted} who specifically target the reconstruction of geometric structure of a human face. Furthermore, recent efforts demonstrate promising results even for in-the-wild scenarios \cite{yu2019inverserendernet,sengupta2018sfsnet}.
Although achieving impressive results relevant to many applications in the fields of computer vision and graphics, these methods were trained to reconstruct the face geometry out of a single image rather than recognizing faces. 
Moreover, in a single image case, a printed image or a tablet screen could be easily used to fool the system. 
As shown, the proposed method can gracefully handle such attacks.

Other methods try to use  CNN with multi-view images to reconstruct the geometry of a scene. 
For example, Yao et al. \cite{yao2018mvsnet,yao2019recurrent} suggested a CNN that uses arbitrary N-view inputs to infer the depth map. 
Other methods \cite{ji2017surfacenet} uses both the images and the camera parameters to enhance the results. 
Still, the above methods do not include specific priors relevant for face recognition. 
In addition, even after successful training of these CNNs, the depth map needs to be further processed in order to be integrated into face-recognition systems (i.e., fed into another CNN to be encoded at latent space). This results in multiple models, expanding the memory footprint and running time of the entire system.

The above methods are all tuned to reconstruct the depth profile of an imaged face rather than recognizing its identity. 
Although the reconstructed surface can play an important role in distinguishing between different individuals, it is still a fundamentally different task.
% that is, the multi-view depth reconstruction. 
%2D-images used for depth inference. 
In this paper, we suggest to use the raw images only, without the need to explicitly estimate the face surface. 
Hence, the CNN  encodes only features that are relevant for the recognition task, utilizing the face geometry in an implicit manner. 
In addition, this results in a single model, trained end-to-end with the raw images, without any need for acquisition or supervision of explicit depth maps during training and inference time.

\subsection{3D Face Recognition} 
Many methods had shown the potential of a given depth map or face surface for face recognition systems. 
Early papers \cite{bronstein2003expression,bronstein2005expression,bronstein2006expression} suggested a model that describes facial expressions as isometric deformations of the facial surface. %
Then, different methods can be used to embed the facial intrinsic geometric structure into a low-dimensional expression-invariant space, used to measure the distance between different face samples. Furthermore, Bronstein et al. \cite{bronstein2004face} suggested a method to embed the isometric deformations without explicitly reconstructing the surface itself.

Other approaches harness 3D morphable models as the embedding model, even in the case of multi-view images, see \cite{tuan2017regressing} as an example. 
In that case, a CNN, or another regression model, estimates the 3DMM's parameters for the input images, and these parameters are used as the embedding vector on which the similarity measure is calculated. 
One of the advantages of using a 3DMM is the disentangling of shape and texture, both of which can be used independently for the recognition task.

Current methods \cite{kim2017deep} also use CNN to process the depth maps directly, and use the deep-features for classification as done in the single-image based methods. 
Here, we use synthetic data with ground truth depth data to show that the proposed method achieves comparable results to that approach, without the need to use the depth maps.  

%____________________________________________________________
\section{Proposed Method}
\subsection{Core Method}
\begin{figure}[htbp]
    \begin{center}
        \includegraphics[width=\linewidth]{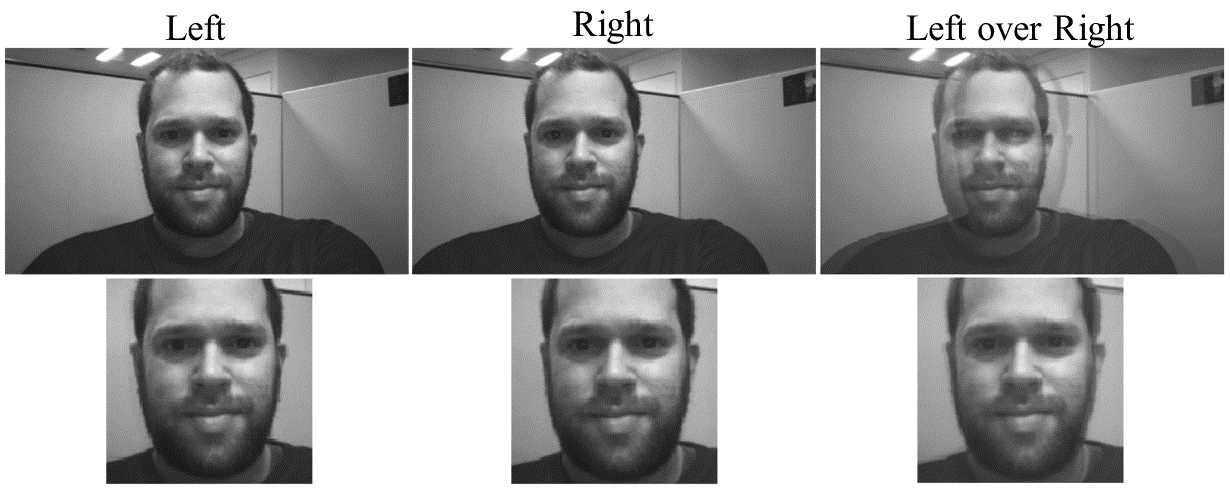}
        \caption{Visualization of the stereo data before and after cropping the face images. 
        In the first row are the original stereo images, in the second row are the cropped images. 
        The third column visualises the differences between the \textit{left} and \textit{right} images. 
        The disparity information is lost if the relative cropping location is not registered as part of the process.}
        \label{fig:origVScropped}
    \end{center}
\end{figure}

In the suggested method we would like to implicitly consider disparity, and hence the geometric surface of the face, as part of the identification phase. 
However, by cropping the face images from the full stereo scene the absolute disparity information is lost as visualized in Figure \ref{fig:origVScropped}. 

\begin{figure}[htb]
    \begin{center}
        \includegraphics[width=\linewidth]{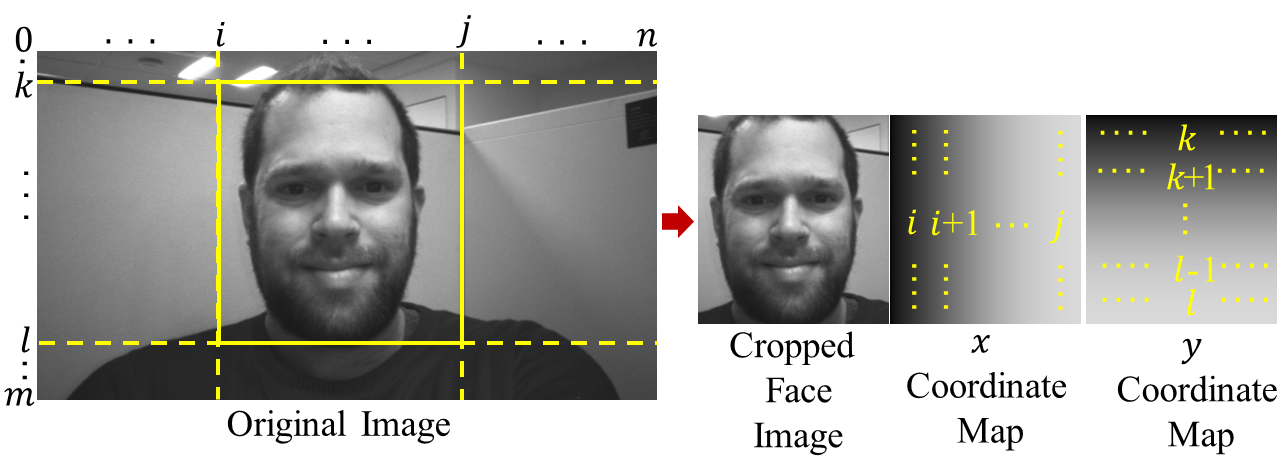}
        \caption{Extracting cropped face image and a matching coordinate map. 
        The values of the pixels in the coordinates map corresponds to the original column and row indices of the matching pixel in the face image.}
        \label{fig:coord_map}
    \end{center}
\end{figure}

In other words, we need to provide the network with the location of the face in each image to maintain the disparity information. 
To that end, we add new channels to each of the cropped face images, containing a mapping of the $x$ and $y$ coordinates of each pixel in the original image before the face was cropped and rescaled (see Figure \ref{fig:coord_map} for a visualization). 
In other words, instead of feeding the CNN with a single input channel (in the case of a gray-scale image), we provide the network with a six-channels input: a stereo pair (\textit{left} and \textit{right}), and the $x$ and $y$ coordinate maps for each image. 
This information is sufficient to extract the geometry of the imaged face.

Since the geometric structure of the face can be used to learn discriminative features, we propose to train the face-recognition network by providing it the identity of the pictured subject without any explicit information about its geometry. 
We assume that the new channels, left image, right image, and the coordinate maps, will promote the encoding of geometric features in the embedding space in an unsupervised manner, that is, without any use of ground-truth depth data. 
As we show in the sequel, the proposed method significantly improves the recognition performances of two state-of-the-art architectures, in comparison to methods that use a single image as an input trained on the same data. 
In addition, we suggest a method to encourage the network to encode meaningful geometric features in order to improve the recognition rates even further. 
Concretely, we use an auxiliary net and loss function to infer geometry-dependent transformations from face-recognition deep features, as described below.

%_________________________________________________________
\subsection{Auxiliary Geometric Net}
\label{section:aux_net}
\begin{figure}[htbp]
    \begin{center}
        \centering
        \includegraphics[width=\linewidth]{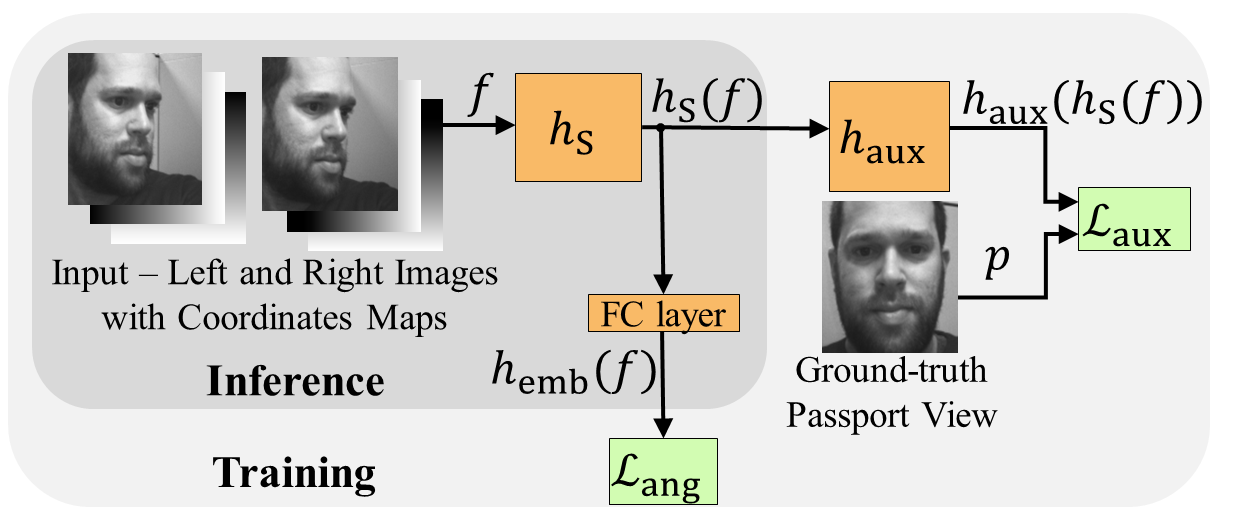}
        \caption{Using auxiliary CNN to account for the geometry. 
        The auxiliary net ($h_\mathrm{aux}$) uses deep features of the core stereo face-recognition CNN ($h_\mathrm{S}$) to estimate a {\it passport} view image, thus, enforcing the core CNN to, at least implicitly, capture the geometry. At inference, only the embedding vector ($h_\mathrm{emb}(f))$ is used for matching.}
        \label{fig:model_flow}
    \end{center}
\end{figure}

To utilize the geometric structure of the face we propose to tune the network towards learning it, at least implicitly. 
An auxiliary CNN model uses deep features extracted from the last convolution block of the core face-recognition network to estimate a frontal {\it passport} view of the input, as can be seen in Figure \ref{fig:model_flow}. 
The motivation for the auxiliary net is the observation that the ``frontalization" (normalization into the frontal pose) of a given face requires an understanding of its geometry. 
%In order to render a fine-detailed image of the face in a different pose, one must have some information regarding the geometry of the scene. 
%
The suggested method does not require explicit estimation or supervision of the geometry. 
Nevertheless, since the auxiliary CNN monitors the deep features of the main face-recognition CNN, it leads those features to include information required for pose normalization. 
In other words, it steers the network to express the geometric structure in its embedding space. 

In order to obtain a good estimate of the \textit{passport} view, and hence more refined geometrical features encoded in the latent space, we use two separate losses. 
The losses require the supervision of ground truth \textit{passport} view image for each subject in the training set. 
We note that an optimal \textit{passport} image will be a frontal image in natural light and a smooth constant background. 
However, the {\it passport} estimation auxiliary task is simple to use also in real data scenarios. 
Given a set of images from a single subject, one can estimate the head pose from the detected landmarks, and choose the most frontal image as the ground truth {\it passport} view image. 
This will cause some variance of small angles in head pose between different \textit{passport} view images, as well as background and illumination variations. 
Nonetheless, we show that this approach is sufficient for increasing the core face recognition performance. 

The first loss term we use is an $\ell_1$ discrepancy between the estimated \textit{passport} view image to the most frontal image of the given subject from the train set used as the ground-truth. 
An additional term is the $\ell_2$ discrepancy between the embedding vector of the estimated \textit{passport} view image produced by a pre-trained single-image (mono) model and the embedding vector of the ground-truth \textit{passport} view image. 
% 
%We denote $h_S$ to be a stereo model and $h_M$ to be a pre-train mono-image model, such that similar inputs to the mono model yields similar embedding vectors. Given a 6-channels stereo input $f_{lr}$ and the matching ground-truth \textit{passport} image $p_{gt}$, we denote the embedding by $h_S(f_{lr})_{emb}$ and the estimated \textit{passport} view as $h_S(f_{lr})_{p}$. 
%The auxiliary loss is given by
The auxiliary loss assumes the form
\begin{eqnarray}
    \mathcal{L}_{\mathrm{aux}} &=& \| h_\mathrm{aux}(h_S(f)) - p \|_1 \cr
    && \,\,\,+ \,\,\alpha \cdot \| h_M(h_\mathrm{aux}(h_S(f))) - h_M(p) \|_2^2;
\end{eqnarray}
where $p$ denotes the ground-truth passport view corresponding to face $f$, $h_S$ and $h_M$ denote the stereo and the mono face embedding models, respectively (we remind that while $h_M$ accepts a single image $p$ as the input, its stereo counterpart $h_S$ operates on the $6$-channel stereo images augmented by the coordinate maps $f$ as previously described), and $h_\mathrm{aux}$ denotes the auxiliary network receiving the deep features and producing an estimated \emph{passport} view.

The total loss of the net is composed of the angular classification loss and the auxiliary one,
$\mathcal{L}  = \mathcal{L}_{\mathrm{ang}} + \beta\cdot \mathcal{L}_{\mathrm{aux}}$ with 
$h_S$ being used as the embedding model in the first term. 
The constants $\alpha$ and $\beta$ control the relative contribution of each loss term in the final objective function. 
The selection of $\beta=0$ yields the core method. 

This approach is general in the sense that it is independent of the face-recognition network architecture. 
Furthermore, the auxiliary net is relevant only during the training process and can be discarded at the inference phase. 
Thus, it does not affect the system's memory and running-time performance. 
Here, we suggest using a ResNet \cite{he2016deep} shaped architecture for the auxiliary CNN, which uses upscale operations at the end of each convolution block.   

%____________________________________________
\section{Experiments and Evaluation}
\subsection{Data}
To the best of our knowledge, there is no public large-scale dataset of stereo face images. 
Therefore, we created our own dataset of real (and synthetic) stereo images.
First, we captured gray-scale images by a standard stereo camera with a baseline of $30mm$, and a resolution of $1920\times1080$ per image. 
The training set includes a total of $344,075$ \textit{left} and \textit{right} images %(a total of $688,150$ single images) 
 of $7,214$ different subjects. 
The test set includes $88,024$ \textit{left} and \textit{right} images of $1,809$ different subjects. 
The images were taken under a variety of lighting conditions and various poses. 
The subjects were located at a distance of less than $1m$ from the camera. %, similar to modern uses in smartphones. 
It is important to mention that the images are not rectified or processed, but rather we use the raw images taken by the stereo camera. 
For the ground truth \textit{passport} images, we chose the most frontal image for each individual subject, as explained in Section \ref{section:aux_net}.

In order to improve our analysis we also generated synthetic data, based on the 3DMM model \cite{blanz1999morphable}. 
We synthesized geometric (and photometric) profiles of $13,473$ different individuals. 
For each subject, we rendered $20-30$ stereo image pairs, with a baseline of $30mm$ between the \textit{left} and \textit{right} images, random pose, random lighting condition, and location in the scene. 
The pose was randomly selected with pitch and yaw angels limited to $\pm 25^o$. 
The distance from the camera was randomly selected within a range of $0.25m-1m$. 
We used random images as a background and added random geometric structures behind the synthetic faces to generate non-flat background information. 
Also, for each image the ground truth depth profile is provided, and for each subject we provide one centered frontal image, that would serve as a reference for our auxiliary {\it passport} view with $pitch=0^o$ and $yaw=0^o$. 
The synthetic training set contains images of $10,103$ distinct subjects and the synthetic test set includes the remaining $3,370$ subjects.

%____________________________________
\subsection{Network Architecture}
As our CNN backbone, we used two state-of-the-art architectures that were introduced and tested in face-recognition challenges, the CosFace \cite{wang2018cosface}, and the ArcFace \cite{deng2019arcface} models. Both architectures use the same ResNet backbone, with different classifiers and angular loss function. 
We compare three methods - our proposed method, depth and texture-based model, and a single image model, denoted as ``stereo", ``depth+texture" and ``mono" models, respectively. 
The depth+texture model receives two channels as input - the face image and the matching ground-truth depth profile, both cropped from the same positions of the original image, and is evaluated only on the synthetic data. For all models, we used 20 ResNet layers with the same number of filters at each layer, as described in \cite{liu2017sphereface}, and an embedding vector of dimension $512$. 
All three models use the same architecture, with the exception that the stereo input involves six channels instead of the mono model's single-channel and the depth+texture model's two channels.  
That affects only the number of parameters in the first convolution layer and is negligible in comparison to the overall size of the network.

%________________________________________________
\subsection{Pre-processing and Data Augmentation}
\label{section:pre_process}
\begin{figure}[htbp]
    \begin{center}
        \centering
        \includegraphics[width=\linewidth]{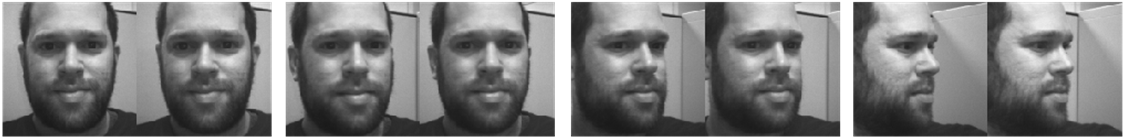}
        \caption{Examples of normalized stereo face images (\textit{left} and \textit{right} images). 
        The stereo model gets both images in two different channels as input. 
        The mono model processes each image as a separate single-channel input.}
        \label{fig:normed_images}
    \end{center}
\end{figure}

We used several common data augmentation techniques. 
For each image, without relation to its stereo pair, we detected face landmarks using automatic facial landmarks detection. 
Then, we cropped a square bound-box around the landmarks, and kept a margin of background around the face. 
Next, we resized the image into a frame size of $144\times144$. Finally, we normalized the gray-scale values of the pixels to be in the range of $[-0.5, 0.5]$. Illustration of normalized stereo images can be seen in Figure \ref{fig:normed_images}. 
As can be seen, as a result of the small baseline of the stereo camera, the images only slightly differ from one another.

We also constructed a matching coordinate map for each cropped frame. Since the coordinates values are in image plane scale (i.e. $1920 \times 1080$), we normalize each column and row index by reducing half of the source image width or height respectively (``zero centering") and dividing by the diagonal length of the source image; given an image of size $W \times H$, the normalized coordinates are given by
\begin{eqnarray}
    (\hat{i}, \hat{j}) &=& \frac{(i-\frac{W}{2}, j-\frac{H}{2})}{\sqrt{W^2+H^2}};
\end{eqnarray}

During training, we cropped a $128\times 128$ sized image in a random location out of the $144\times 144$ face image and its corresponding coordinate map. This augmentation purpose is to be robust to the error in the landmark detector, which affects the resulting cropped face image. Consequently, The input dimensions are $128\times 128\times 6$ for the stereo model, $128\times128\times 2$ for the depth+texture model, and $128\times128\times 1$ for the single image model. Also, we added random noise and blur augmentation to the texture channels.
Finally, for testing, we cropped the center $128\times 128$ bounding box of the normalized input, and did not add noise to the input data.

%_________________________________________
\subsection{Training}
All three models were trained with the same hyper-parameters, except for the batch size, as will be explained shortly, for the same amount of epochs, and using the same data. 
While the stereo model was trained based on a couple of \textit{left}-\textit{right} images, the mono model was trained with both the \textit{left} and \textit{right} images treated independently. 
Thus, the batch size used for training the mono model was twice as large as that of the stereo, compensating for the fact each sample of the stereo model involves two images. 
For training the auxiliary net, we used $\alpha=50,\beta=1$, as loss normalization factors. 

We used Stochastic Gradient Descent optimizer to train all of the models for $100$ epochs, with an initial learning rate of $0.01$ and reduced the learning rate by a factor of $0.1$ each $20$ epochs. We used a batch size of $64$ for the mono model and $32$ for the stereo model, and a weight decay factor of $0.0005$.%

%_____________________________________
\subsection{Evaluation and Results}
\subsubsection{Experimental Results on Synthetic Data}
First, we trained and evaluated our system using the synthetic dataset, for which we have access to ground truth depth profiles and to the illumination parameters used to render the images. 
We explored the robustness of the different methods that deal with head pose and illumination variations, as two of the most fundamental attributes of depth-based recognition systems. 
We compared our method to the baseline single-image based model and to a CNN explicitly operating on both the face image and the ground-truth depth profiles given as inputs. 
We look at the positive samples, that is, two samples of the same subject, and record the similarity scores together with the head pitch and yaw angles of both samples. 

\begin{figure}[htb]
    \begin{center}
        \centering
        \includegraphics[width=\linewidth]{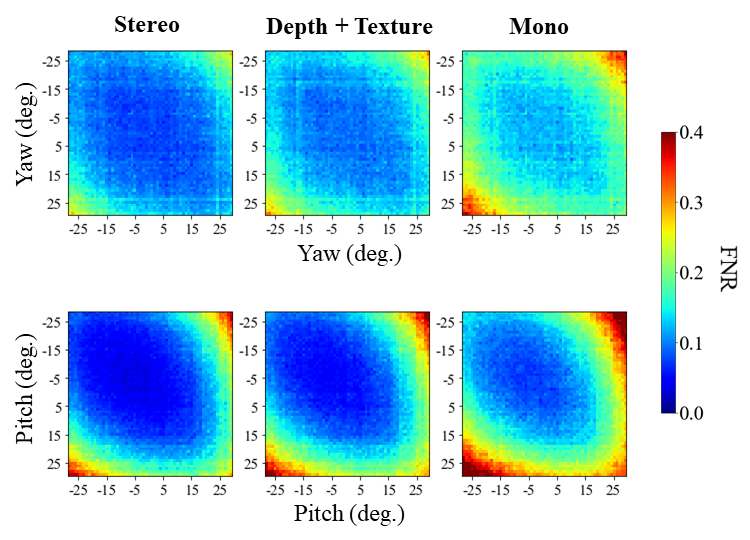}
        \caption{FNR of positive couples for the stereo, depth, and mono approaches as a function of the variations in the yaw (first row) and pitch (second row) angles. Warmer colors represent a higher FNR (hence a larger number of false miss-identifications).}
        \label{fig:FNR_per_angle_analysis}
    \end{center}
\end{figure}

Next, we set a threshold for each of the models, that achieves a false-positive rate (FPR) of $10^{-6}$. 
Then, we measured the relative false-negative rate (FNR) per yaw and pitch angles, that is, which fraction of the samples with a given yaw and pitch angles got similarity score smaller than the threshold. 
The results are shown in Figure \ref{fig:FNR_per_angle_analysis}. 
The stereo model is more robust to both high pitch and high yaw angles compared to the mono model, and attains performance level on par with the depth-based model. 
The features produced by the stereo model are much more robust to head pose variability than those of the mono model, a similar behavior to the depth-based features, with the exception that the stereo model achieves this without using any explicit depth information. 

\begin{figure}[htb]
    \begin{center}
        \centering
        \includegraphics[width=\linewidth]{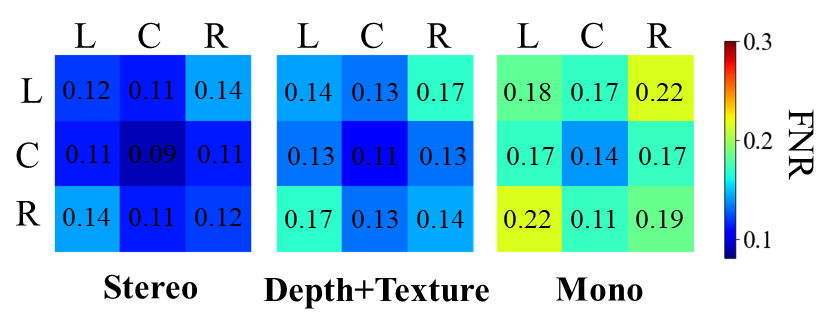}
        \caption{FNR of positive couples for the stereo, depth, and mono approaches as a function of the variations in the illumination conditions: left directional light, direct center light, and right directional light (denoted as L,C,R respectively.)}
        \label{fig:FNR_per_light}
    \end{center}
\end{figure}

We repeated this experiment and compared the robustness to illumination variation. To that end, we separated the illumination of each image to one of three types: left directional light, direct center light, or right directional light. We measured the FNR for samples with different combinations of these lighting types, as displayed in Figure \ref{fig:FNR_per_light}. Again, the stereo model achieves the best performance, in particular for the case of opposite light directions across samples. 

\begin{figure}[hbt!]
    \begin{center}
        \centering
        \includegraphics[height=4cm]{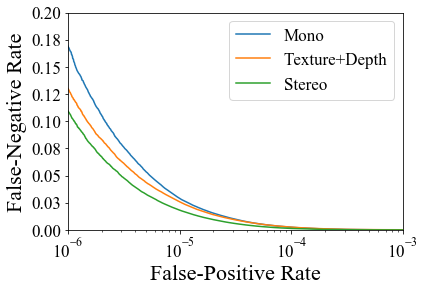}

        \caption{ROC curve on the synthetic images dataset.}
        \label{fig:left+depth_ROC_curve}
    \end{center}
\end{figure}

Finally, we measured the recognition rates for the different methods. 
%note: the following explanation is copied from the real data experiment section
For each model, we set thresholds that enabled us to obtain certain false-positive rate (FPR) for the negative samples, that is, different subjects that were identified as the same subject. 
For each threshold and FPR, we measured the false-negative rate (FNR) of the positive pairs, that is, samples of the same subject that were identified as two different subjects. 
In order to compare the same samples in all models, in the mono model the embedding vectors of the \textit{left} and \textit{right} images pair were combined into a single vector.
We tested several methods: concatenating the vectors, averaging, and choosing only the left or only the right image vector. 
Averaging the embedding vectors achieved the best performance, and this is the method displayed in the results below.
The ROC curve of this experiment is displayed in Figure \ref{fig:left+depth_ROC_curve}.
The suggested method achieves better performance for all thresholds. 
Lower FPR represents a higher threshold needed for discrimination between different subjects, which in turn translates to higher FNR. 
Since the stereo model is much more robust to big head pose and illumination variation, it suffers smaller degradation when using a higher threshold, similarly to the depth-based model. 
While the difference between the stereo model and the mono model increases as the FPR get's smaller, the difference between the stereo and the depth-based models is stable after FPR$=10^-5$, suggesting they achieve similar behavior.
Although we do not claim the method used to process the explicit depth profiles is the optimal one, the experiment does give us quantitative results to support the claim that the proposed method is at least comparable to using explicit depth profiles as an input, and achieves the same properties without using any explicit depth information.

%_____________________________________
\subsubsection{Experimental Results on Real Data}
In addition to the evaluation of the synthetic dataset, we trained and tested our method on real stereo images.
A total of $1,635,336$ subjects pairs have been evaluated, with $2,288,368$ positive samples of the same subject at different positions, poses, and illumination conditions, and $472,612,104$ negative samples, of two different subjects.

\begin{figure}[htbp]
    \begin{center}
        \centering
        \includegraphics[height=4cm]{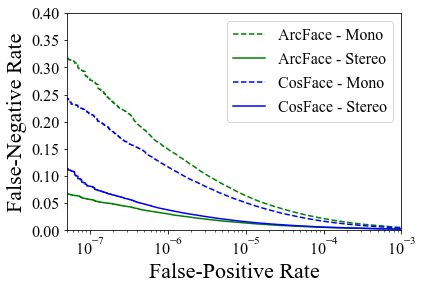}
        \caption{ROC curve for the real images dataset. 
        The dashed lines correspond to the mono models, while the solid lines correspond to the stereo counterparts.}
        \label{fig:ROC_curve}
    \end{center}
\end{figure}

The results are presented in Table 1. %\ref{table:results}. 
The methods are denoted as either {\it Mono} or {\it Stereo}. 
In addition, a full ROC-curve is displayed in Figure \ref{fig:ROC_curve}. 
In both tested architectures, the stereo model significantly outperforms the mono model.

\begin{table}[htb]
\begin{center}
\label{table:results}
\begin{tabular}{lllll}
%\noalign{\smallskip}
\hline
\hline
\noalign{\smallskip}
%\hline\noalign{\smallskip}
Model & FPR: & $10^{-5}$ & $2\cdot10^{-6}$ & $10^{-6}$\\
\noalign{\smallskip}
\hline
\noalign{\smallskip}
CosFace - Mono && 0.0567 & 0.1048 & 0.1329\\
\textbf{CosFace - Stereo} && \textbf{0.0161} & \textbf{0.0297} & \textbf{0.0378}\\
\hline
ArcFace - Mono && 0.0640 & 0.1211 & 0.1496\\
\textbf{ArcFace - Stereo} && \textbf{0.0135} & \textbf{0.0243} & \textbf{0.0304}\\
\hline
\end{tabular}
\end{center}
\caption{FNR at different FPR on the real images dataset, for Stereo and Mono methods implemented with different models.}
\end{table}

%_____________________________
\subsubsection{Ablation Study}
We performed multiple experiments to study the effect of the suggested method and auxiliary task on real data. 
We used the CosFace classifier to test different variants of the proposed methods. 
First, we trained a stereo model without supplying the coordinate maps as part of the input. 
That is, the input was only a stereo image pair, with dimensions $128\times 128\times 2$, denoted as {\it No Coords}. 
We also trained a model that uses the coordinate maps, without using the auxiliary net, denoted as {\it With Coords}. 
Then, we trained a model with the auxiliary net but with just the $\ell_1$ loss ($\alpha=0, \beta=1$), and finally, the complete method ($\alpha=50, \beta=1$), both denoted as \textit{With Aux}. 
The values for $\alpha,\beta$ were chosen empirically.
The results are shown in Table 2. %\ref{table:ablation_study}. 
\begin{table*}[htb]
\begin{center}
\label{table:ablation_study}
\begin{tabular}{lllll}
\hline
\hline
\noalign{\smallskip}
Stereo Method \hfill & FPR: & $10^{-5}$ & $2\cdot10^{-6}$ & $10^{-6}$\\
\noalign{\smallskip}
\hline
\noalign{\smallskip}
No Coords && 0.0254 & 0.0479 & 0.0628\\
With Coords && 0.0235 & 0.0407 & 0.0503\\
With Aux ($\alpha=0,\beta=1$)& & 0.0216 & 0.03823 & 0.04719\\
\textbf{With Aux ($\alpha=50,\beta=1$)} && \textbf{0.0161} & \textbf{0.0297} & \textbf{0.0378}\\
\hline
\end{tabular}
\end{center}
\caption{FNR at different FPR on the real images dataset using different components of the suggested method}
\end{table*}

Even without the coordinate maps the stereo model outperforms the mono model. 
This means, that the joint learning of slightly different viewpoints allows the net to learn more distinctive features in comparison to the processing of each image independently. 
Nonetheless, adding the coordinate maps channels significantly enhanced the results. 
These channels contain only the geometric location of the head in the scene, without any additional data regarding the face texture or other conditions in the scene such as pose or lighting conditions. 
Thus, we conclude that the net uses the coordinate maps to encode more distinctive features in the latent space, possibly by computing the geometric relations between some facial features. 

\begin{figure}[pt]
    \begin{center}
        \centering
        \includegraphics[width=\linewidth]{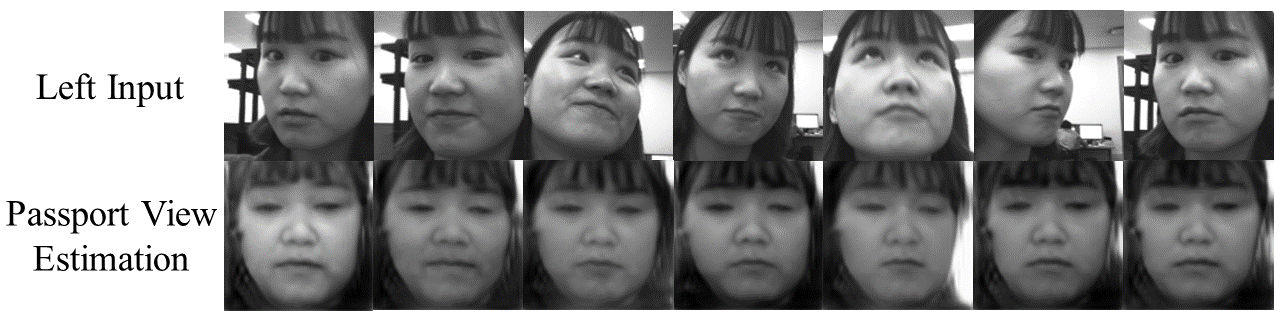}
        \caption{Results of the auxiliary CNN, estimating the passport view of a given face. 
        The input to the auxiliary net is the deep features from the core face recognition CNN.
        } 
        \label{fig:passport_estimation}
    \end{center}
\end{figure}

Training using the auxiliary net also demonstrates a significant improvement in performance. 
Using both the suggested auxiliary losses provided the best results. Once again, the auxiliary net uses only the deep features to estimate a new viewpoint of the scene, without any additional input. 
This supports our assumption that the suggested auxiliary task promotes learning of geometric features in the latent space, and thus leads to better recognition rates. 
A visualisation of the estimated \textit{passport} views is presented in Figure \ref{fig:passport_estimation}.

%________________________________

\subsubsection{Anti-Spoofing}
This section demonstrates the resiliency to spoofing attacks.
We detect attempts to hack the authentication system using images of a given face, displayed on a screen or printed on a paper. 
Such cases pose a great challenge for single-image based methods. 
The difference between the geometric structures of a flat image and a human face can be utilized to overcome this problem without explicit reconstruction of the geometric structure. 
As training data, we simulated flat printout versions of some of our faces.
By taking an original face image, assigning each pixel a point coordinate in 3D space $(x,y,z)$, and projecting it onto a stereo camera system (\textit{left} and \textit{right} images), we simulate a flat printout of the image. 
The projection to the stereo cameras is done using a pinhole camera model. 
Given a point $(x,y,z)$ in world coordinates, the camera focal point $f$, and the camera baseline $b$, the projection of the point to the camera system coordinates $(x_p,y_p)$ is calculated by
%}
\begin{eqnarray}
    \binom{x_p}{y_p} &=&
            \frac{f}{z} \cdot \binom{x - b}{y}.
\end{eqnarray}
For each image, we assigned an arbitrary depth ($z$ coordinates) which simulates flat plane profile of natural size, with random pitch and yaw angles in range $0-20$ $deg$.
Then, we project it onto the stereo system, and use it as a spoofing attack example. 
We used the same pre-processing and augmentation techniques  describe in Section \ref{section:pre_process}, and trained a classification network to classify each input to either real (label=$1$) or a spoofing attack (label=$0$). 
For evaluation, we used the following spoofing attacks.
Printed images on A4 papers, TV and laptop screens displaying images of new subjects that were not part of the training. For each stereo images pair, we predicted whether it is a real person, or a spoofing attack.
We evaluate the model over $510$ stereo images of printed faces, 446 stereo images of screens, and 3,580 stereo images of real people. 
We note that the training data consists only of projections of the real face images. 
Hence, it does not include 2D features which commonly exist in real-world spoofing attacks such as the texture of the material, edges of the screen or page, reflections of light on the screens surface, etc.
Still, the stereo model detected $99.8\%$ of the papers and $96.2\%$ of the screens images, while kipping correct classification of $97.2\%$ of the real persons examples. Thus, we suggest that the model utilized geometric features to discriminate the attacks from the real persons.
%}
%_______________________________________________

\section{Conclusions}
We presented a novel method to encode geometric facial features in a face-recognition CNN model, without the need to explicitly capture or calculate geometrical data. 
We showed that our method outperforms traditional single-image based methods, and is comparable to methods that use explicit and accurate depth data. We also showed that our method can utilize geometric features to detect spoofing attacks.
Thus, we bridge the gap between the advantages of 3D face recognition methods and the simplicity and cost-efficiency of existing 2D face recognition methods.

The proposed method is simple to implement, can be used with any general CNN architecture, and is robust to both classic face-recognition challenges such as extreme poses, and geometric data acquisition and calibration distortions. 
We demonstrated the potential of using the raw data from multiple view scenarios to improve computational face authentication rates and robustness, without the need to explicitly reconstruct the geometry. Future studies can expand this method to additional applications that currently require explicit depth information such as hand-gesture recognition and body pose estimation.

\section*{Acknowledgments}
This research was partially supported by the Israel Ministry of Science and Technology grant number 3-14719.

{\small
\bibliographystyle{ieee}
\bibliography{egbib}

\begin{thebibliography}{10}\itemsep=-1pt

\bibitem{blanz1999morphable}
V.~Blanz and T.~Vetter.
\newblock A morphable model for the synthesis of 3d faces.
\newblock In {\em Proceedings of the 26th annual conference on Computer
  graphics and interactive techniques}, pages 187--194, 1999.

\bibitem{blanz2003face}
V.~Blanz and T.~Vetter.
\newblock Face recognition based on fitting a 3d morphable model.
\newblock {\em IEEE Transactions on pattern analysis and machine intelligence},
  25(9):1063--1074, 2003.

\bibitem{bronstein2003expression}
A.~M. Bronstein, M.~M. Bronstein, and R.~Kimmel.
\newblock Expression-invariant 3d face recognition.
\newblock In {\em international conference on Audio-and video-based biometric
  person authentication}, pages 62--70. Springer, 2003.

\bibitem{bronstein2005expression}
A.~M. Bronstein, M.~M. Bronstein, and R.~Kimmel.
\newblock Expression-invariant face recognition via spherical embedding.
\newblock In {\em IEEE International Conference on Image Processing 2005},
  volume~3, pages III--756. IEEE, 2005.

\bibitem{bronstein2006expression}
A.~M. Bronstein, M.~M. Bronstein, and R.~Kimmel.
\newblock Expression-invariant representations of faces.
\newblock {\em IEEE Transactions on Image Processing}, 16(1):188--197, 2006.

\bibitem{bronstein2004face}
A.~M. Bronstein, M.~M. Bronstein, A.~Spira, and R.~Kimmel.
\newblock Face recognition from facial surface metric.
\newblock In {\em European Conference on Computer Vision}, pages 225--237.
  Springer, 2004.

\bibitem{deng2019arcface}
J.~Deng, J.~Guo, N.~Xue, and S.~Zafeiriou.
\newblock Arcface: Additive angular margin loss for deep face recognition.
\newblock In {\em Proceedings of the IEEE Conference on Computer Vision and
  Pattern Recognition}, pages 4690--4699, 2019.

\bibitem{erdogmus2013spoofing}
N.~Erdogmus and S.~Marcel.
\newblock Spoofing 2d face recognition systems with 3d masks.
\newblock In {\em 2013 International Conference of the BIOSIG Special Interest
  Group (BIOSIG)}, pages 1--8. IEEE, 2013.

\bibitem{he2016deep}
K.~He, X.~Zhang, S.~Ren, and J.~Sun.
\newblock Deep residual learning for image recognition.
\newblock In {\em Proceedings of the IEEE conference on computer vision and
  pattern recognition}, pages 770--778, 2016.

\bibitem{ji2017surfacenet}
M.~Ji, J.~Gall, H.~Zheng, Y.~Liu, and L.~Fang.
\newblock Surfacenet: An end-to-end 3d neural network for multiview stereopsis.
\newblock In {\em Proceedings of the IEEE International Conference on Computer
  Vision}, pages 2307--2315, 2017.

\bibitem{kim2017deep}
D.~Kim, M.~Hernandez, J.~Choi, and G.~Medioni.
\newblock Deep 3d face identification.
\newblock In {\em 2017 IEEE international joint conference on biometrics
  (IJCB)}, pages 133--142. IEEE, 2017.

\bibitem{liu2017sphereface}
W.~Liu, Y.~Wen, Z.~Yu, M.~Li, B.~Raj, and L.~Song.
\newblock Sphereface: Deep hypersphere embedding for face recognition.
\newblock In {\em Proceedings of the IEEE conference on computer vision and
  pattern recognition}, pages 212--220, 2017.

\bibitem{paysan20093d}
P.~Paysan, R.~Knothe, B.~Amberg, S.~Romdhani, and T.~Vetter.
\newblock A 3d face model for pose and illumination invariant face recognition.
\newblock In {\em 2009 Sixth IEEE International Conference on Advanced Video
  and Signal Based Surveillance}, pages 296--301. Ieee, 2009.

\bibitem{richardson20163d}
E.~Richardson, M.~Sela, and R.~Kimmel.
\newblock 3d face reconstruction by learning from synthetic data.
\newblock In {\em 2016 Fourth International Conference on 3D Vision (3DV)},
  pages 460--469. IEEE, 2016.

\bibitem{richardson2017learning}
E.~Richardson, M.~Sela, R.~Or-El, and R.~Kimmel.
\newblock Learning detailed face reconstruction from a single image.
\newblock In {\em Proceedings of the IEEE Conference on Computer Vision and
  Pattern Recognition}, pages 1259--1268, 2017.

\bibitem{sela2017unrestricted}
M.~Sela, E.~Richardson, and R.~Kimmel.
\newblock Unrestricted facial geometry reconstruction using image-to-image
  translation.
\newblock In {\em Proceedings of the IEEE International Conference on Computer
  Vision}, pages 1576--1585, 2017.

\bibitem{sengupta2018sfsnet}
S.~Sengupta, A.~Kanazawa, C.~D. Castillo, and D.~W. Jacobs.
\newblock Sfsnet: Learning shape, reflectance and illuminance of facesin the
  wild'.
\newblock In {\em Proceedings of the IEEE Conference on Computer Vision and
  Pattern Recognition}, pages 6296--6305, 2018.

\bibitem{singh2018techniques}
S.~Singh and S.~Prasad.
\newblock Techniques and challenges of face recognition: A critical review.
\newblock {\em Procedia computer science}, 143:536--543, 2018.

\bibitem{tuan2017regressing}
A.~Tuan~Tran, T.~Hassner, I.~Masi, and G.~Medioni.
\newblock Regressing robust and discriminative 3d morphable models with a very
  deep neural network.
\newblock In {\em Proceedings of the IEEE conference on computer vision and
  pattern recognition}, pages 5163--5172, 2017.

\bibitem{wang2018cosface}
H.~Wang, Y.~Wang, Z.~Zhou, X.~Ji, D.~Gong, J.~Zhou, Z.~Li, and W.~Liu.
\newblock Cosface: Large margin cosine loss for deep face recognition.
\newblock In {\em Proceedings of the IEEE Conference on Computer Vision and
  Pattern Recognition}, pages 5265--5274, 2018.

\bibitem{yao2018mvsnet}
Y.~Yao, Z.~Luo, S.~Li, T.~Fang, and L.~Quan.
\newblock Mvsnet: Depth inference for unstructured multi-view stereo.
\newblock In {\em Proceedings of the European Conference on Computer Vision
  (ECCV)}, pages 767--783, 2018.

\bibitem{yao2019recurrent}
Y.~Yao, Z.~Luo, S.~Li, T.~Shen, T.~Fang, and L.~Quan.
\newblock Recurrent mvsnet for high-resolution multi-view stereo depth
  inference.
\newblock In {\em Proceedings of the IEEE Conference on Computer Vision and
  Pattern Recognition}, pages 5525--5534, 2019.

\bibitem{yu2019inverserendernet}
Y.~Yu and W.~A. Smith.
\newblock Inverserendernet: Learning single image inverse rendering.
\newblock In {\em Proceedings of the IEEE Conference on Computer Vision and
  Pattern Recognition}, pages 3155--3164, 2019.

\bibitem{zhang2016joint}
K.~Zhang, Z.~Zhang, Z.~Li, and Y.~Qiao.
\newblock Joint face detection and alignment using multitask cascaded
  convolutional networks.
\newblock {\em IEEE Signal Processing Letters}, 23(10):1499--1503, 2016.

\end{thebibliography}
}

\end{document}